\documentclass[10pt,twocolumn,letterpaper]{article}

\usepackage{cvpr}
\usepackage{color}
\usepackage{times}
\usepackage{epsfig}
\usepackage{graphicx}
\usepackage{amsmath}
\usepackage{amssymb}

\definecolor{accolornotes}{rgb}{0.7,0.3,0.2}

\definecolor{changes}{rgb}{0.45,0,0}

\definecolor{jscolornotes}{rgb}{0.3,0.7,0.2}

\definecolor{acgray}{rgb}{0.8,0.8,0.8}

\definecolor{myred}{rgb}{0.6, 0, 0}
\definecolor{myblue}{rgb}{0.3, 0.1, 0.9}

\newcommand{\be}{\begin{equation}}
\newcommand{\ee}{\end{equation}}
\newcommand{\bea}{\begin{eqnarray}}
\newcommand{\eea}{\end{eqnarray}}
\newcommand{\beas}{\begin{eqnarray*}}
\newcommand{\eeas}{\end{eqnarray*}}

\renewcommand{\eqref}[2][\reflabel]{(\ref{eq:#1-#2})} 

\newcommand{\reflabel}{dummy} 

\def\vec#1{\mathchoice{\mbox{\boldmath  $\displaystyle\bf#1$}}
{\mbox{\boldmath  $\textstyle\bf#1$}}
{\mbox{\boldmath  $\scriptstyle\bf#1$}}
{\mbox{\boldmath  $\scriptscriptstyle\bf#1$}}}

\def\mat#1{\mathchoice{\mbox{\boldmath$\displaystyle\tt#1$}}
{\mbox{\boldmath$\textstyle\tt#1$}}
{\mbox{\boldmath$\scriptstyle\tt#1$}}
{\mbox{\boldmath$\scriptscriptstyle\tt#1$}}}

\DeclareMathAlphabet{\mathcal}{OMS}{cmsy}{m}{n}

\usepackage[pagebackref=true,breaklinks=true,letterpaper=true,colorlinks,bookmarks=false]{hyperref}

\cvprfinalcopy 

\def\cvprPaperID{2015-58} 

\ifcvprfinal\fi
\begin{document}

\title{Decision Forests, Convolutional Networks and the Models in-Between\\{\large Microsoft Research Technical Report 2015-58}}

\author{Yani Ioannou$^1$ \and Duncan Robertson$^2$ \and Darko Zikic$^2$ \and Peter Kontschieder$^2$ \and Jamie Shotton$^2$ \and Matthew Brown$^3$ \and Antonio Criminisi$^2$ \and
$^1$University of Cambridge,
$^2$Microsoft Research,
$^3$University of Bath
}

\maketitle

\begin{abstract}
This paper investigates the connections between two state of the art classifiers: 
decision forests (DFs, including decision jungles) and convolutional neural networks (CNNs).
Decision forests are computationally efficient thanks to their {\em conditional computation} property (computation is confined to only a small region of the tree, the nodes along a single branch).
CNNs achieve state of the art accuracy, thanks to their {\em representation learning} capabilities.
\\
We present a systematic analysis of how to fuse conditional computation with representation learning and achieve a continuum of hybrid models with different ratios of accuracy \vs efficiency. 
We call this new family of hybrid models {\em conditional networks}. 
Conditional networks can be thought of as: 
i) decision trees augmented with data transformation operators, or
ii) CNNs, with block-diagonal sparse weight matrices, and explicit data routing functions.\\
Experimental validation is performed on the common task of image classification on both the CIFAR and Imagenet datasets. 
Compared to state of the art CNNs, our hybrid models yield the same accuracy
with a fraction of the compute cost and much smaller number of parameters.
\end{abstract}

\section{Introduction}

Machine learning has enjoyed much success in recent years for both academic and commercial scenarios.
Two learning approaches have gained particular attention: 
(i) random forests~\cite{amit97shape,Breiman2001,Criminisi2013book,Shotton2013}, as used \eg in Microsoft Kinect~\cite{Kinect}; and 
(ii) deep neural networks (DNNs)~\cite{Krizhevsky2012imanet,Sermanet2013overfeat}, as used for speech recognition~\cite{yu2014book} and image classification~\cite{He2015delving}, among other applications.
Decision trees are characterized by a \emph{routed} behavior: conditioned on some learned routing function, the data is sent either to one child or another.  
This \emph{conditional computation} means that at test time only a small fraction of all the nodes are visited, thus achieving high efficiency.
Convolutional neural networks repeatedly transform their input through several (learned) non-linear transformations. Typically, at each layer all units  need to perform computation.
CNNs achieve state-of-the-art accuracy in many tasks, but decision trees have the potential to be more efficient. This paper investigates the connections between these two popular models, highlighting differences and similarities in theory and practice. 

\noindent{\bf Related work.}
Decision forests were introduced in~\cite{amit97shape,Breiman2001} as efficient models for classification and regression. Forests were extended to density estimation, manifold learning and semi-supervised learning in~\cite{Criminisi2013book}. 
The decision jungle variant~\cite{Shotton2013} replaces trees with DAGs (directed acyclical graphs) 
to reduce memory consumption.

Convolutional networks were introduced for the task of digit recognition in~\cite{leCun1990ConvNets}. More recently they have been applied with great success to the task of image classification over 1,000 classes~\cite{Ba2013dothey,Denil2013predicting,Denton2014efficient,
He2015delving,Krizhevsky2012imanet,Lin2013NiN,Sermanet2013overfeat,
Simonyan2014verydeep,Szegedy2014going,Zhang2015efficient}.

In general, decision trees and neural networks are perceived to be very different models. 
However, the work in~\cite{Sethi1990,Welbl2014casting} demonstrates how any decision tree or DAG can be represented as a two-layer perceptron with a special pattern of sparsity in the weight matrices. 
Some recent papers have addressed the issue of mixing properties of trees and convolutional networks together.
For example,  the two-routed CNN architecture in~\cite{Krizhevsky2012imanet} is a stump (a tree with only two branches). GoogLeNet~\cite{Szegedy2014going} is another example of a (imbalanced) tree-like CNN architecture.

The work in~\cite{Sun2013cascade,Toshev2014deeppose} combines multiple ``expert'' CNNs into one, manually designed DAG architecture. 
Each component CNN is trained on a specific task (\eg detecting an object part), using a part-specific loss. 
In contrast, here we investigate training a single, tree-shaped CNN model by minimizing one global training loss. 
In our model the various branches are not explicitly trained to recognize parts (though they may do so if this minimizes the overall loss).

The work in~\cite{Zheng2014pedestrian} is a cascade~\cite{viola2004robust} of CNN classifiers, each trained at a different level of recognition difficulty. Their model does not consider tree-based architectures.
Finally, the work in~\cite{Kontschieder2015DNDF} achieves state of the art classification accuracy by replacing the fully-connected layers of a CNN with a forest. This model is at least as expensive as the original CNN since the convolutional layers (where most of the computation is) are not split into different branches.

\noindent{\bf Contributions.}
The contributions of this paper are as follows:
i) We show how DAG-based CNN architectures (namely {\em conditional networks}) with a rich hierarchical structure (\eg high number of branches, more balanced trees) produce classification accuracy which is at par with state of the art, but with much lower compute and memory requirements. 
ii) We demonstrate how conditional networks are still differentiable despite the presence of explicit data routing functions. 
iii) We show how conditional networks can be used to fuse the output of CNN ensembles in a data driven way, yielding higher accuracy for fixed compute. 
Validation is run on the task of image-level classification, on both the CIFAR and Imagenet datasets.

\section{Structured Sparsity and Data Routing}
\label{sec:background}

The seminal work in~\cite{Krizhevsky2012imanet} demonstrated how 
introducing rectified linear unit activations (ReLUs) allows {\em deep} CNNs to be trained effectively. 
Given a scalar input $v_j$, its ReLU activation is $\sigma(v_j)=\max(0, v_j)$. 
Thus, this type of non-linearity \emph{switches off} a large number of feature responses within a CNN.
ReLU activations induce a data-dependent sparsity; but this sparsity does not tend to have much structure in it. Enforcing a special type of {\em structured} sparsity is at the basis of the efficiency gain attained by conditional networks. We illustrate this concept with a toy example.

\begin{figure}[t]
\centerline{
\includegraphics[width=1.05\linewidth]{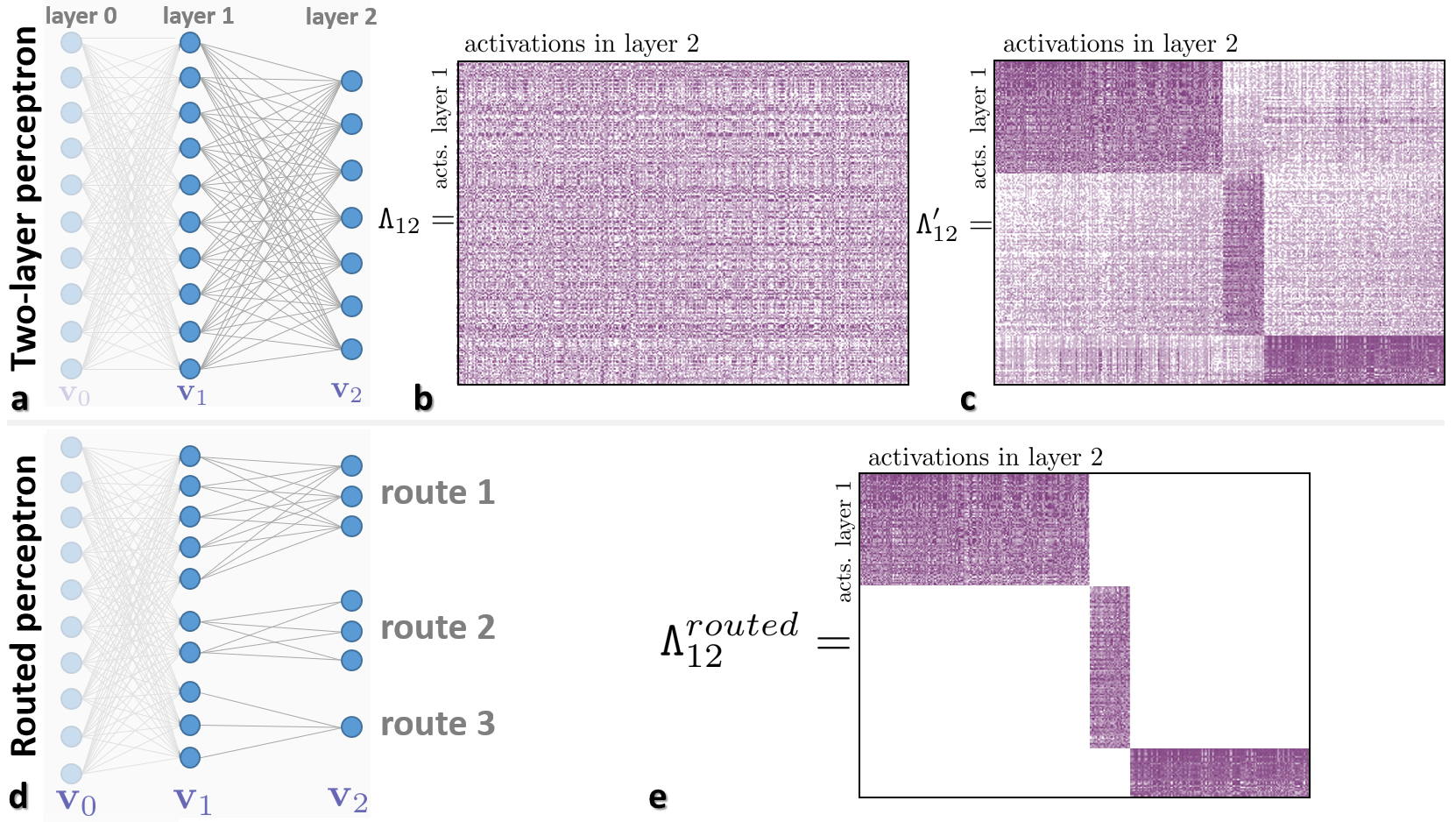}
}
   \caption{{\bf Block-diagonal correlation of activations, and data routing.}
   {\bf (a)} An example 2-layer preceptron with ReLU activations. This is a portion of the `VGG' model~\cite{Simonyan2014verydeep} trained on Imagenet.
   {\bf (b)} The correlation matrix $\mat \Lambda_{12}$ shows {\em unstructured} activation correlation between unit pairs. 
   {\bf (c)} Reordering the units reveals a noisy, block-diagonal structure. 
   {\bf (e)} Zeroing-out the off-diagonal elements is equivalent to removing connections between unit pairs. This corrsponds to the sparser, {\em routed} perceptron in {\bf (d)}.
   }
\label{fig:sparsity}
\end{figure}

The output of the exemplar multi-layer perceptron (MLP) of  Fig.~\ref{fig:sparsity}a is computed as $\vec v_2 = \sigma(\mat P_{12} \vec v_1) = \sigma(\mat P_{12} \sigma(\mat P_{01} \vec v_0))$.
Given a trained MLP we can look at the average correlation of activations between 
pairs of units in two successive layers, over all training data. 
For example, the matrix $\mat \Lambda_{12}$ (Fig.~\ref{fig:sparsity}b) shows the joint correlations 
of activations in layers 1 and 2 in a perceptron trained on the Imagenet classification task.\footnote{The correlation matrix $\mat \Lambda_{12}$ is not the same as the weight matrix $\mat P_{12}$.} 
Here we use the final two layers of the deep CNN model of~\cite{Simonyan2014verydeep} with 
a reduced number of features (250) and classes (350) to aid visualization. 

Thanks to the ReLUs, the correlation matrix $\mat \Lambda_{12}$ has many zero-valued elements (in white in Fig.~\ref{fig:sparsity}b), and these are distributed in an {\em unstructured} way. 
Reordering the rows and columns of $\Lambda_{12}$ reveals an underlying, noisy block-diagonal pattern  (Fig.~\ref{fig:sparsity}c). 
This operation corresponds to finding groups of layer-1 features which are highly active for certain subsets of classes (indexed in layer-2).
Thus, the darker blocks in Fig.~\ref{fig:sparsity}c correspond to three super-classes (sets of `related' classes). 
Zeroing out the off-diagonal elements (Fig.~\ref{fig:sparsity}e) corresponds to removing connections between corresponding unit pairs. 
This yields the sparse architecture in Fig.~\ref{fig:sparsity}d, where selected subsets of the layer-1 features are sent (after transformation) to the corresponding subsets of layer-2 units; thus giving rise to data routing.

We have shown how imposing a block-diagonal pattern of sparsity to the joint activation correlation 
in a neural network corresponds to equipping the network with a tree-like, 
routed architecture. Next section will formalize this intuition further and show the benefits of 
sparse architectures.

\section{Conditional Networks: Trees or Nets?}
\label{sec:model}

This section introduces the conditional networks model, 
in comparison to trees and CNNs, and discusses their efficiency and training.
For clarity, we first introduce a compact graphical notation for representating both 
trees and CNNs.

\begin{figure}[t]
\centerline{
\includegraphics[width=0.9\linewidth]{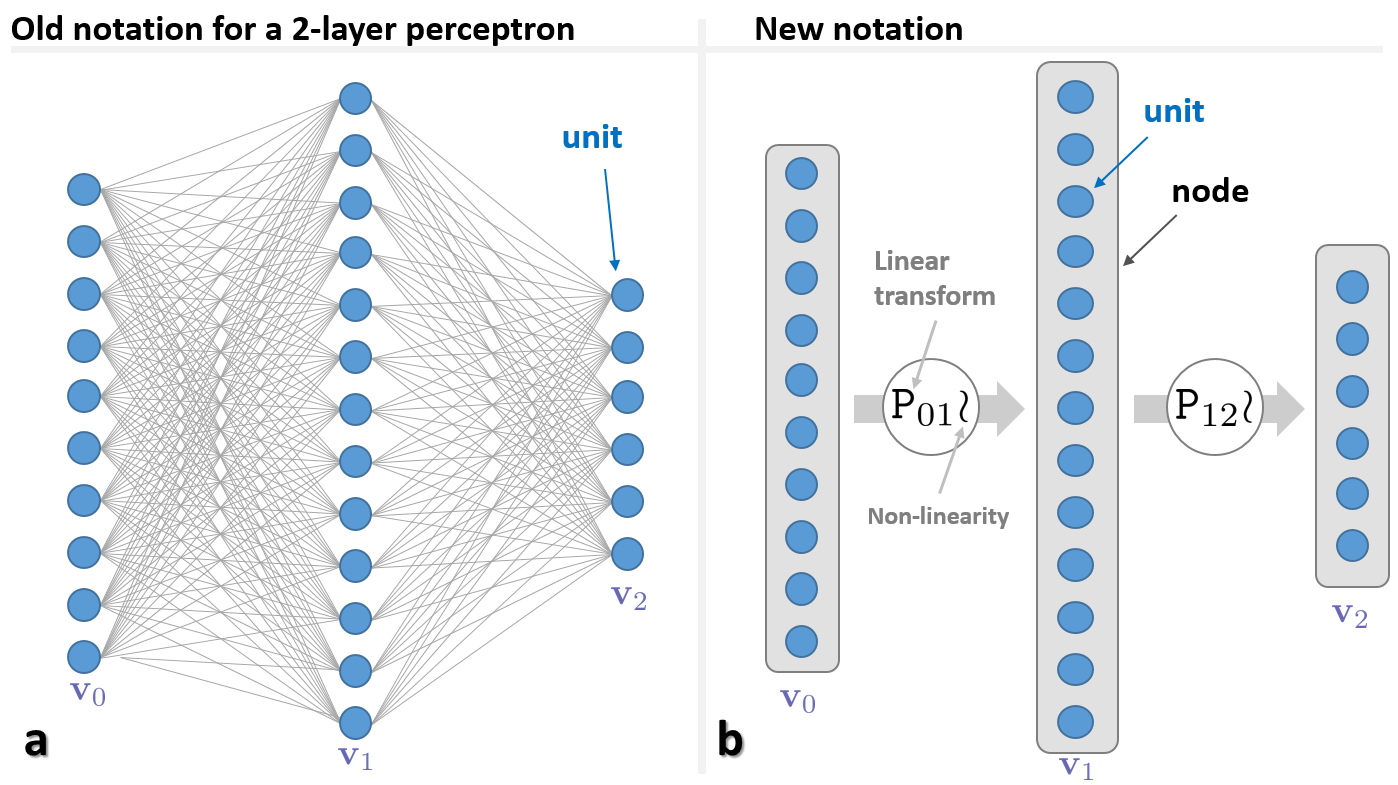}
}
   \caption{{\bf A compact graphical notation for neural networks}. Data transformation is indicated by the projection matrix $\mat P$ followed by a non-linearity (denoted with the symbol $\wr$). The bias term is not shown here as we use homogeneous coordinates.}
\label{fig:newGraphLanguage}
\end{figure}

{\bf Representing CNNs.}
Figure~\ref{fig:newGraphLanguage}a shows the conventional way to represent an MLP, with its units (circles) connected via edges (for weights).
Our new notation is shown in Fig.~\ref{fig:newGraphLanguage}b, where the symbol 
$\mat P_{ij} \wr$ denotes the popular non-linear transformation $\vec v_j = \sigma(\mat P \vec v_i)$
between two consecutive layers $i$ and $j$. 
The linear projection matrix is denoted `$\mat P$', and `$\wr$'
indicates a non-linear function $\sigma(.)$ (\eg a sigmoid or ReLU). 
In the case of CNNs the function $\sigma$ could also incorporate \eg max-pooling and drop-out. 
Deep CNNs are long concatenations of the structure in Fig.~\ref{fig:newGraphLanguage}b. 


\begin{figure}[t]
\centerline{
\includegraphics[width=0.75\linewidth]{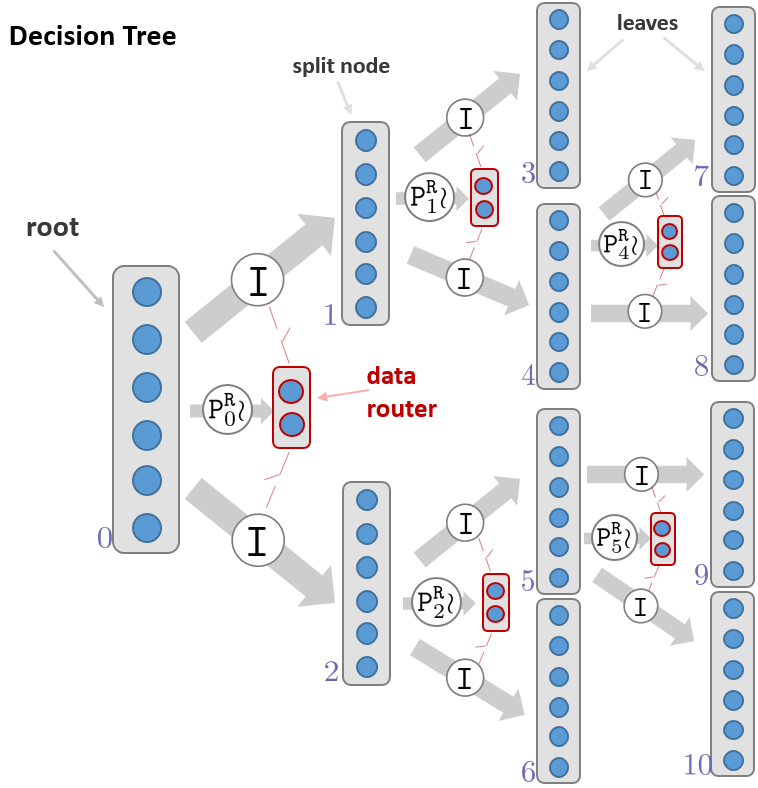}
}
   \caption{{\bf Representing decision trees}. 
   Data routing functions (a.k.a. routers, in red) direct the data to one or more child nodes.
   Identity matrices copy the data without transforming it.
}
\label{fig:decisionTree}
\end{figure}

{\bf Representing trees and DAGs.}
The same graphical language can also represent trees and DAGs (Fig.~\ref{fig:decisionTree}). 
Usually, in a tree the data is moved from one node to another untransformed.\footnote{This is in contrast to representation learning approaches which estimate optimal data transformation processes. Exceptions are~\cite{montillo2011entangled,BuloKontsch2014}.}
In our notation this is achieved via the identity matrix $\mat I$
(\ie $\vec v_j = \mat I \vec v_i$).
Additionally, {\em Selecting} a subset of features $\vec v'$ from a longer vector $\vec v$ is achieved 
as $\vec v' = \mat S \vec v$, with $\mat S$  non-square matrix with only one element per row equal to 1, and 0 everywhere else. Identity and selection transforms are special instances of linear projections.

A key operator of trees which is not present in CNNs is data routing.
Routers send the incoming data to a selected sub-branch and enable conditional computation.
Routers (red nodes in Fig.~\ref{fig:decisionTree}) are represented here as perceptrons, 
though other choices are possible.
In general, a router outputs {\em real-valued} weights, which may be used to select 
a single best route, multiple routes ({\em multi-way routing}), 
or send the data fractionally to all children ({\em soft routing}).

A conditional network exhibits both data routing and non-linear data transformation
within a highly branched architecture
(Fig.~\ref{fig:conditionalNetwork}).

\subsection{Computational Efficiency}
\label{sec:efficiency}

\begin{figure}[t]
\centerline{
\includegraphics[width=0.9\linewidth]{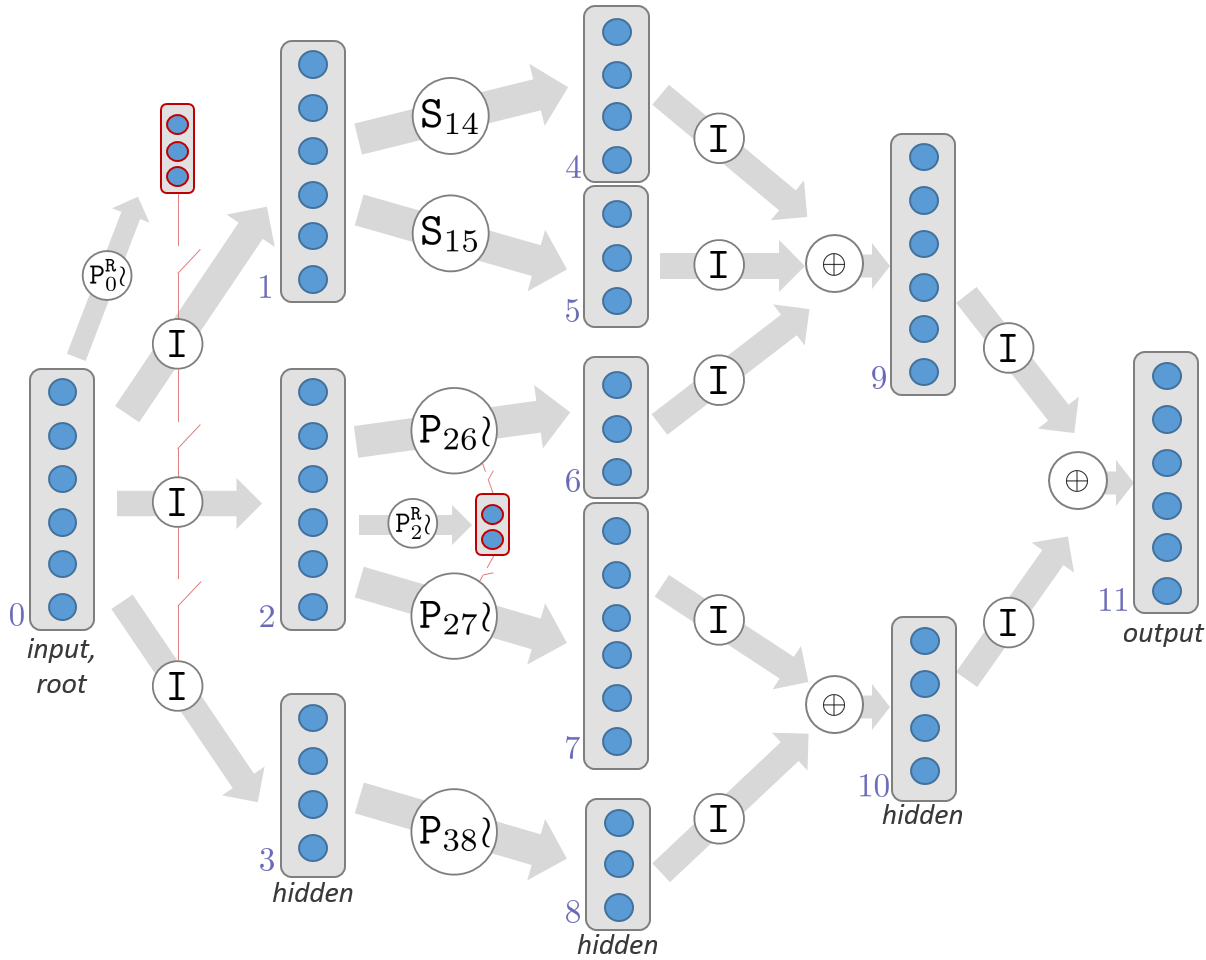}
}
   \caption{{\bf A generic conditional network.} 
   Conditional networks fuse efficient data routing with accurate data transformation in a single model. 
   Vector concatenations are denoted with $\oplus$.}
\label{fig:conditionalNetwork}
\end{figure}

\noindent{\bf Efficiency through explicit data routing.}
Split nodes can have {\em explicit} routers where data is {\em conditionally} sent to the children according to the output of a routing function (\eg node 2 in Fig.~\ref{fig:conditionalNetwork}),
or have \emph{implicit} routers where the data is unconditionally but selectively sent to the children using selection matrices $\mat S$ (\eg node 1).
If the routing is explicit and hard (like in trees), then successive operations will be applied to ever smaller subsets of incoming data, with the associated compute savings. 
Next we show how implicit conditional networks can also yield efficiency. 

\begin{figure}[t]
\centerline{
\includegraphics[width=1.15\linewidth]{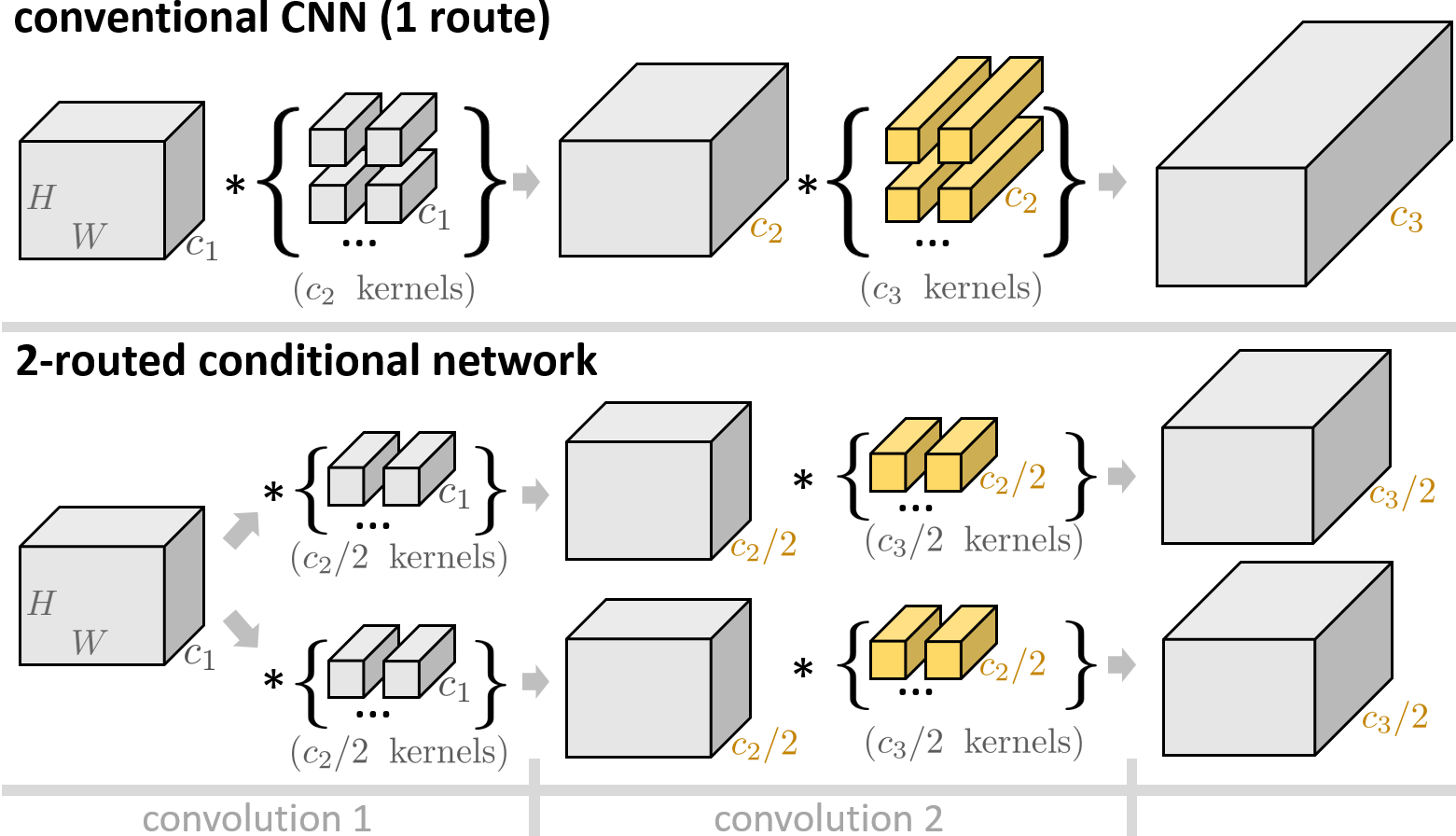}
}
\caption{{\bf Computational efficiency of implicit conditional networks.} 
{\bf (top)} A standard CNN (one route).
{\bf (bottom)} A two-routed architecture with no explicit routers.
The larger boxes denote feature maps, the smaller ones the filters.
Due to branching, the depth of the second set of kernels (in yellow) changes between the two architectures.
The reduction in kernel size yields fewer computations and thus higher efficiency in the branched network.
}
\label{fig:implicitEfficiency}
\end{figure}

\noindent{\bf Efficiency of implicit routed networks.}
Figures~\ref{fig:implicitEfficiency} compares a standard CNN with a 2-routed
architecture. The total numbers of filters at each layer is fixed for 
both to $c_1$, $c_2$ and $c_3$.
The number of multiplications necessary in the first convolution is 
$c_2 \times c_1 k_x k_y W H$, with $W, H$ the size of the feature 
map and $k_x, k_y$ the kernel size (for simplicity here we ignore max-pooling operations).
This is the same for both architectures.
However, due to routing, the depth of the second set of filters is different between the two architectures.
Therefore, for the conventional CNN the cost of the second convolution is 
$c_3 \times c_2 k_x k_y W H$, while for the branched architecture 
the cost is $c_3 \times \left( \frac{c_2}{2}\right) k_x k_y W H$, \ie half the cost of the standard CNN. 
The increased efficiency is due {\em only} to the fact that shallower kernels are convolved with shallower feature maps. 
Simultaneous processing of parallel routes may yield additional 
time savings.\footnote{Feature not yet implemented in Caffe~\cite{Toshev2014deeppose}.}

\subsection{Back-propagation Training}
\label{sec:training}
\begin{figure}[t]
\centerline{
\includegraphics[width=0.6\linewidth]{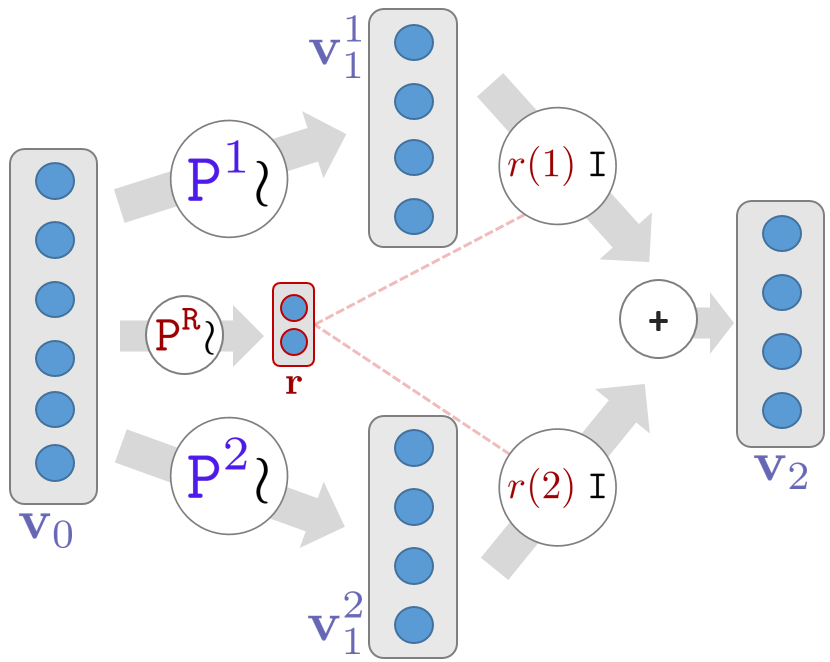}
}
   \caption{{\bf Training a network's routers via back-propagation.} 
   A toy conditional network used to illustrate how to train the router's parameters $\mat P^{\tt R}$ via gradient descent back-propagation.}
\label{fig:training}
\end{figure}

{\em Implicitly}-routed conditional networks can be trained with the standard back-propagation algorithm~\cite{Krizhevsky2012imanet,Szegedy2014going}. 
The selection functions $\mat S$ become extra parameters to optimize over, and their gradients can be derived straightforwardly. 
Now we show that {\em explicitly}-routed networks can also be trained using back-propagation.  
To do so we need to compute partial derivatives with respect to the router's parameters (all other differentiation operations are as in conventional CNNs). 
We illustrate this using the small network in Fig.~\ref{fig:training}. 
Here subscripts index layers and superscripts index routes (instead, in Fig.~\ref{fig:conditionalNetwork} the subscripts indexed the input and output nodes).
The training loss to be minimized is
\begin{equation}
L(\boldsymbol\theta)  = \frac{1}{2} \left(\vec v_2(\boldsymbol\theta) - \vec v_2^*\right)^\top \left(\vec v_2(\boldsymbol\theta) - \vec v_2^*\right),
\end{equation}
with $\boldsymbol\theta = \{ \{ \mat P^j \}, \mat P^{\tt R} \} $ denoting the parameters of the network, and $\vec v_2^*$ the ground-truth assignments to the output units. We define this energy for a single training data point, though the extension to a full dataset is a trivial outer summation.
The network's forward mapping is
\begin{equation}
\vec v_1^j = \sigma \left( \mat P^j \vec v_0 \right) \;\;\;\; \textrm{and} \;\;\;\; \vec v_2(\boldsymbol\theta) = \vec r(\boldsymbol\theta) \; \mat V_1(\boldsymbol\theta),
\end{equation} 
with $\vec r= \sigma \left( \mat P^{\tt R} \vec v_0 \right)$ the output of the router.
In general: 
i) the routing weights $\vec r$ are {\em continuous}, $r(i)\in[0,1]$, and
ii) multiple routes can be ``on'' at the same time.
$\mat V_1$ is a matrix whose $j$-th row is $(\vec v_1^j) ^\top$.
The update rule is
$
\Delta \boldsymbol\theta_{t+1}  := -\rho \left. \frac{\partial E}{\partial \boldsymbol\theta} \right|_t,
$ with $t$ indexing iterations. We compute the partial derivatives through the chain rule as follows:
\begin{equation}
\frac{\partial L}{\partial \boldsymbol\theta} = 
\frac{\partial L}{\partial \vec v_2} \; 
\frac{\partial \vec v_2}{\partial \boldsymbol\theta} = 
\frac{\partial L}{\partial \vec v_2} \left( 
\frac{\partial \textcolor{myred}{\vec r}}{\partial \textcolor{myred}{\mat P^{\tt R}}} \mat V_1 +  
\sum_{j=1}^{R} \;\textcolor{myred}{r(j)}\; 
\frac{\partial \vec v_1^j}{\partial \boldsymbol\phi^j} 
\frac{\partial \boldsymbol\phi^j}{\partial \textcolor{myblue}{\mat P^j}}\right)
\label{eq:chainrule},
\end{equation}
with $\boldsymbol{\phi}^j := \mat P^j \vec v_0$, and $R$ the number of routes. 
Equation~(\ref{eq:chainrule}) shows the influence of the soft routing weights on the 
back-propagated gradients, for each route.
Thus, explicit routers can be trained as part of the overall back-propagation procedure. 
Since trees and DAGs are special instances of conditional networks, 
now we have a recipe for training them via back-propagation (\cf~\cite{Kontschieder2015DNDF,Schulter2013Alternating,Suarez1999GlobalTree}).

In summary, conditional networks may be thought of as:
i) Decision trees/DAGs which have been enriched with (learned) data transformation operations, or as
ii) CNNs with rich, DAG-shaped architectures and trainable data routing functions. 
Next, we show efficiency advantages of such branched models with comparative experiments.

\section{Experiments and Comparisons}
\label{sec:experiments}

Conditional networks generalize decision trees, DAGs and CNNs, and thus
could be used in all tasks where those are successful. 
Here we compare those models with one another on the popular task of image-level classification. 
We explore the effect of different ``branched'' architectures 
on a joint measure of: i) classification accuracy, ii) test-time compute cost and iii) model size.

\subsection{Conditional Sparsification of a Perceptron}
\label{sec:results_perceptron}
\begin{figure}[t]
\centerline{
\includegraphics[width=1.04\linewidth]{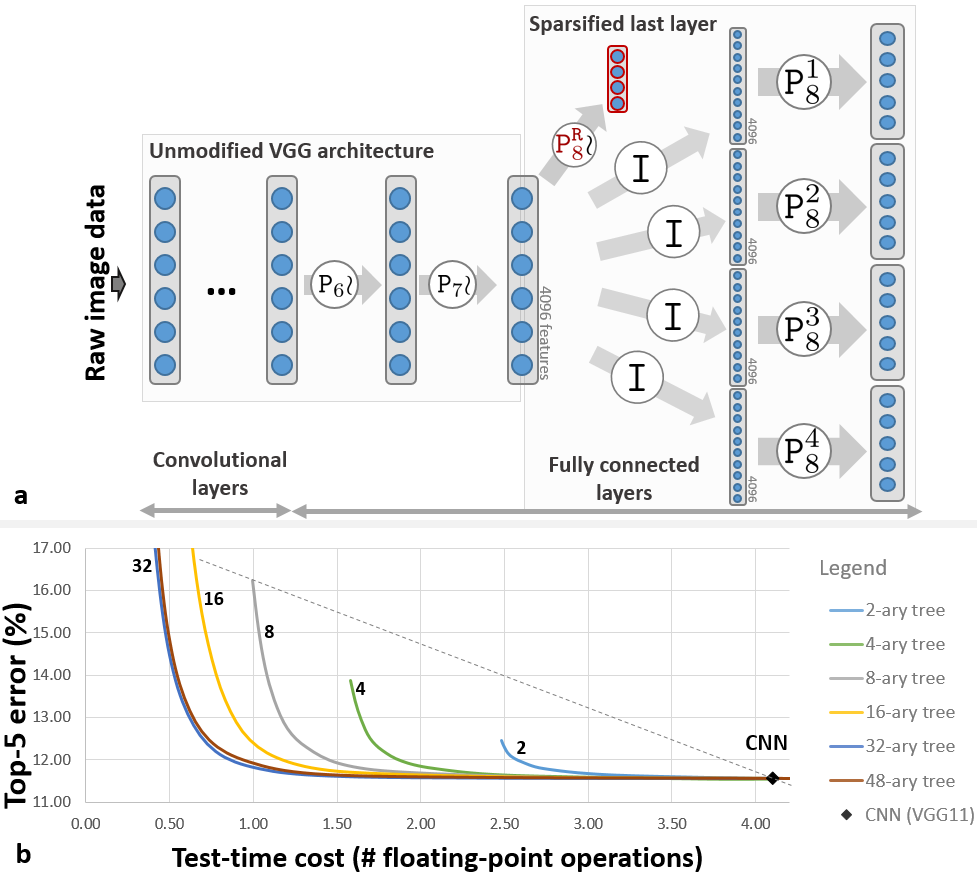}
}
   \caption{{\bf Conditional sparsification of a single-layer perceptron.}
   {\bf (a)}  We take the deep CNN model in~\cite{Simonyan2014verydeep} (`VGG11') and turn the last fully connected layer (size $4095\times1000$) into a tree with $R$ routes ($R=4$ shown in figure).
   {\bf (b)} The top-5-error \vs test-time-cost curves for six conditional networks trained with different values of $R\in\{2,4,6,8,16,24,32\}$. 
Test-time cost is computed as number of floating point operations per image, and is hardware-independent.
The strong sub-linear shape of the curves indicates a net gain in the trade-off between accuracy and efficiency. 
   }
\label{fig:resultsSinglePerc}
\end{figure}

We begin with a toy experiment, designed to illustrate potential advandages of using explicit routes within a neural network. We take a perceptron (the last layer of ``VGG11''~\cite{Simonyan2014verydeep}) and
train it on the 1,000 Imagenet classes, with no scale or relighting augmentation~\cite{Jia2014caffe}.
Then we turn the perceptron into a small tree, with $R$ routes and an additional, compact perceptron as a router (see fig.~\ref{fig:resultsSinglePerc}a). The router ${\mat P}^{\tt R}_8$ and the projection matrices ${\mat P}^i_8$ are trained to minimize the overall classification loss (Sec.~\ref{sec:training}).

\noindent{\bf Interpolating between trees and CNNs.}
Given a test image we apply the convolutional layers until the beginning of the tree. 
Then we apply the router, and its $R$ outputs are soft-max normalized and treated 
as probabilities for deciding which route/s to send the image to. 
We can send the image only to the highest probability route only (as done in trees)
or we could send it to multiple routes, \eg the $\tau$ most probable ones. 
For $\tau=1$ we reproduce the behaviour of a tree. 
This corresponds to the left-most point in the curves in 
fig.~\ref{fig:resultsSinglePerc}b (lowest cost and higher error). 
Setting $\tau=R$ corresponds to sending the image to {\em all} routes. 
The latter reproduces the same behaviour as the CNN, with nearly the same 
cost (lowest error and highest compute cost point in the curves).
Different values of $\tau \in \{1,\dots,R\}$ correspond to different points along the error-cost curves.

\noindent{\bf Dynamic accuracy-efficiency trade-off.}
The ability to select the desired accuracy-efficiency operating point 
at {\em run-time} allows \eg better battery management in mobile applications.
In contrast, a CNN corresponds to a {\em single} point in the accuracy-efficiency space 
(see the black point in fig.~\ref{fig:resultsSinglePerc}b). 
The pronounced sub-linear behaviour of the curves in fig.~\ref{fig:resultsSinglePerc}b 
suggests that we can increase the efficiency considerably with little accuracy reduction
(in the figure a 4-fold efficiency increase yields an increase in error of less than $1\%$). 

\noindent{\bf Why care about the amount of computation?}
Modern parallel architectures (such as GPUs) yield
high classification accuracy in little time. But parallelism is not the only way of 
increasing efficiency. Here we focus on reducing the total amount of
computations while maintaining high accuracy. 
Computation affects power consumption, which is of huge practical 
importance in mobile applications (to increase battery life on a smartphone) 
as well as in cloud services (the biggest costs in data centres are due to their cooling).
Next we extend conditional processing also to the expensive convolutional layers of a deep CNN.

\subsection{Comparing Various Architectures on Imagenet}

Here we validate the use of conditional networks for image classification in the ILSVRC2012 dataset~\cite{ILSVRC2015}. 
The dataset consists of $\sim$1.2M training images for 1000 classes, and 50K validation images.
We base our experiments on the VGG network~\cite{Simonyan2014verydeep} on which the 
current best models are also based~\cite{He2015delving}. 
Specifically, we focus on the VGG11 model as it is deep (11 layers) and relatively memory efficient 
(trains with Caffe~\cite{Jia2014caffe} on a single Nvidia K40 GPU).

\noindent{\bf Global max-pooling.}
We found that using {\em global} max-pooling, 
after the last convolutional layer is effective in reducing the number of parameters while maintaining the same accuracy.
We trained a new network (`VGG11-GMP') with such pooling, and achieved lower top-5 error than the baseline VGG11 
network (13.3\% \vs 13.8\%), with a decrease in the number of parameters of over 72\%.

\begin{figure}[t]
\centerline{
\includegraphics[width=1.1\linewidth]{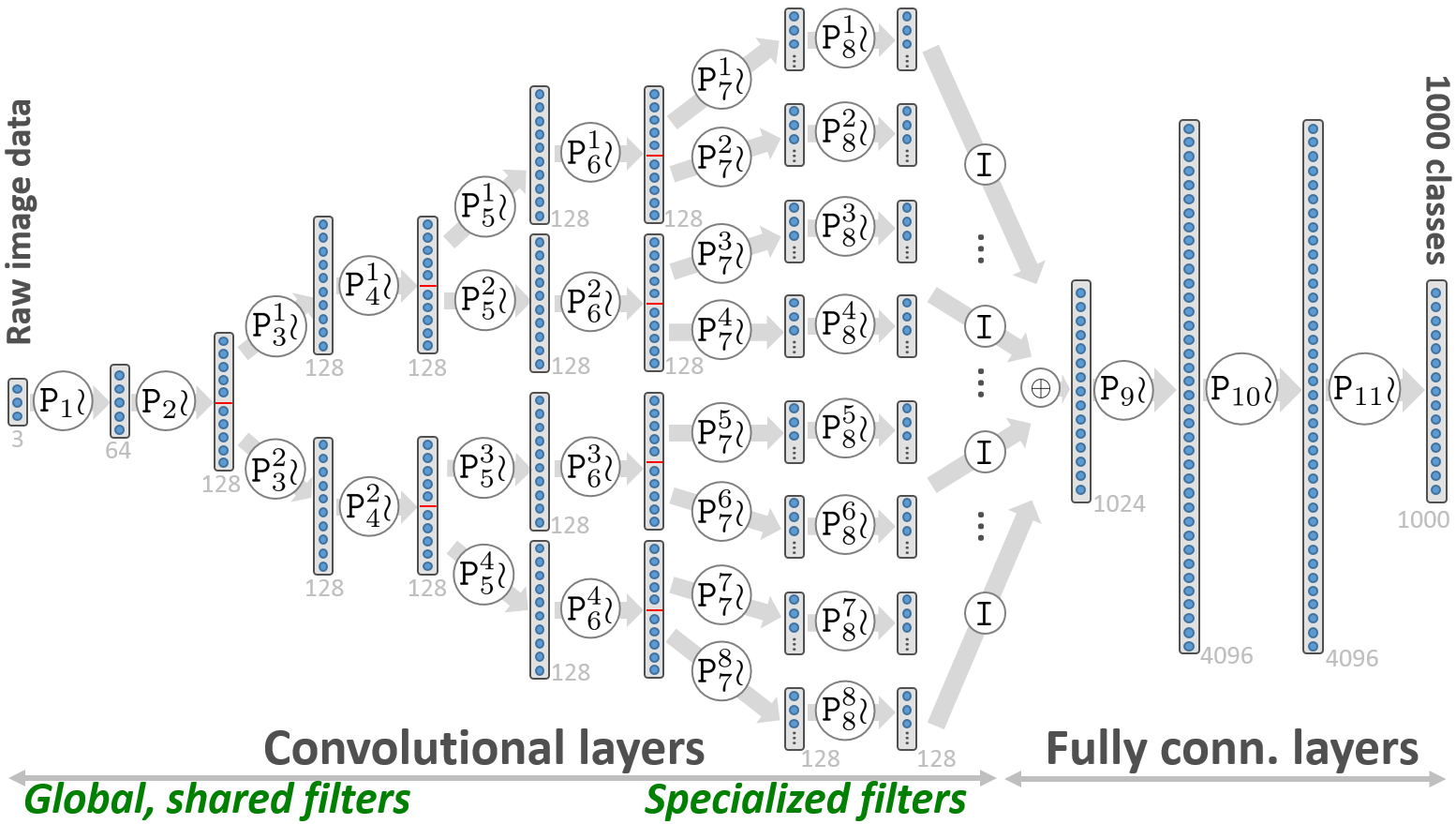}
}
   \caption{{\bf The conditional network used on the Imagenet experiments} employs implicit data routing in the 
   (usually expensive) convolutional layers to yield higher compute efficiency than 
   the corresponding, unrouted deep CNN (here VGG11). Small red lines indicate ``groups'' of feature maps as implemented in Caffe.
}
\label{fig:Imagenet_CondNet}
\end{figure}

\noindent{\bf Designing an efficient conditional architecture.}
Then we designed the conditional network in Fig.~\ref{fig:Imagenet_CondNet} by starting with the
unrouted VGG11-GMP and splitting the convolutional layers (the most computationally expensive layers) into a DAG-like, routed architecture. The hypothesis is that each filter should only need to be applied to a small number of channels in the input feature map.
%
Data routing is implemented via {\em filter groups}~\cite{Krizhevsky2012imanet}. 
Thus, at the $n$-th convolutional level (with $n=3\ldots 5$)
the filters of VGG11-GMP are divided into $2^{(n-2)}$ groups. 
Each group depends only on the results of 128 previous filters. 
The feature maps of the last convolutional layer are concatenated together, and globally max-pooled
 before the single-routed, fully-connected layers, which remain the same as those in VGG11-GMP.

\begin{figure*}[t]
\centerline{
\begin{tabular}{ccc}
\includegraphics[width=0.37\linewidth]{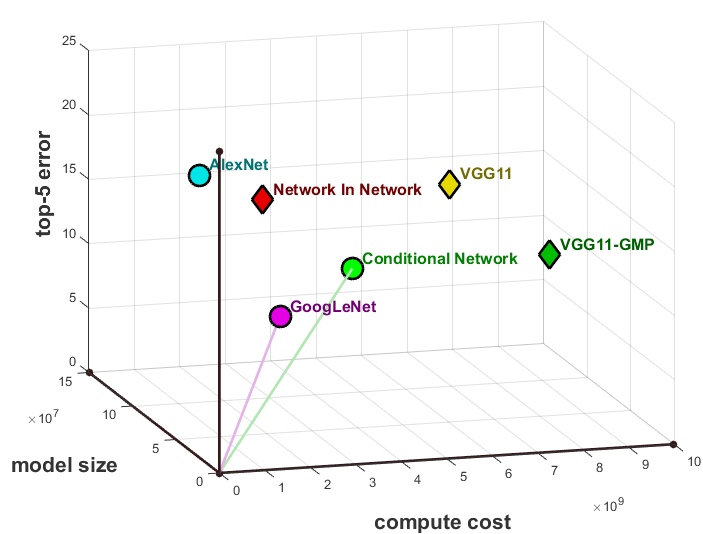}&\hspace{-0.6cm}
\includegraphics[width=0.31\linewidth]{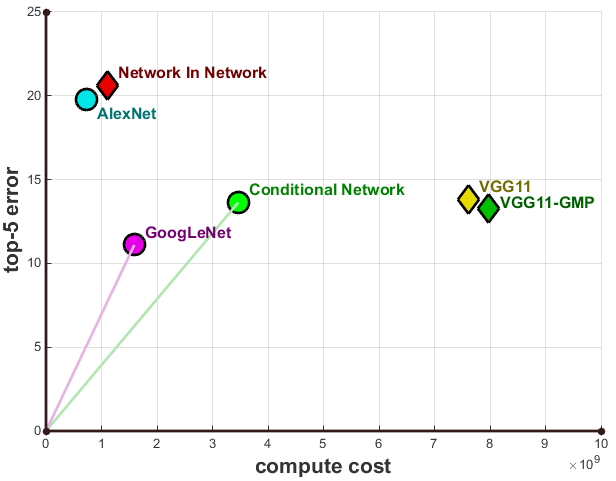}&\hspace{-0.6cm}
\includegraphics[width=0.31\linewidth]{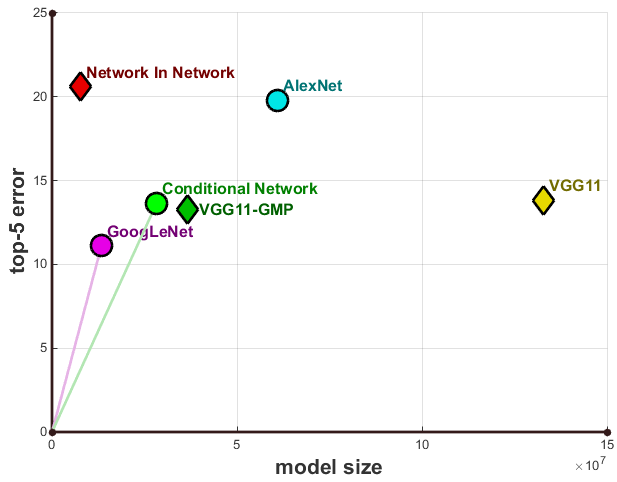}\\
\end{tabular}
}
\caption{{\bf Comparing different network architectures on Imagenet.} 
Top-5 error as a function of test-time compute and model size, for various networks, 
validated on the Imagenet dataset.
{\bf (left)} A 3D view.
{\bf (middle)} Error \vs compute cost.
{\bf (right)} Error \vs model size.
Our VGG11-GMP net (dark green) reduces model size significantly. 
Conditional networks (denoted with circles) yield points closest to the origin, corresponding to the best accuracy-efficiency trade-off. 
The conditional architecture of fig.\ref{fig:Imagenet_CondNet} is the second closest to the origin.
}
\label{fig:Imagenet_results}
\end{figure*}

\noindent{\bf Training.}
We trained the architecture in Fig.~\ref{fig:Imagenet_CondNet} {\em from scratch}, 
with the same parameters as in~\cite{Simonyan2014verydeep}, 
except for using the initialization of~\cite{He2015delving}, and a learning schedule of 
$\gamma_t = \gamma_0(1+\gamma_0\lambda t)^{-1}$, where $\gamma_0,\gamma_t$ and $\lambda$ 
are the initial learning rate, learning rate at iteration $t$, and weight decay, respectively~\cite{Bottou2012sgdtricks}. 
When the validation accuracy levelled out, the learning rate was decreased by a factor 10, twice. 
Our architecture took twice as many epochs to train than VGG11, but thanks to higher 
efficiency it took roughly the same time.

\noindent{\bf Results: accuracy \vs compute \vs size.}
In order to compare different network architectures as fairly as possible, here we did not use any training
 augmentation aside from that supported by Caffe (mirroring and random crops). 
 Similarly, we report test-time accuracy based only on centre-cropped images, without potentially expensive data oversampling. 
 This reduces the overall accuracy (w.r.t.\ to state of the art), but constitutes a fairer test bed for 
 teasing out the effects of different architectures.
Applying the same oversampling to all networks produced a similar accuracy improvement in 
all models, without changing their ranking.

Figure~\ref{fig:Imagenet_results} shows top-5 error as a function of test-time compute cost and model size,
for various architectures.
Compute cost is measured as the number of multiply-accumulate operations. 
We chose this measure of efficiency because it is directly related to the theoretical complexity
on the testing (run-time) algorithm, and it is machine/implementation independent.
Later we will also show how in our parallel implementation this measure of efficiency correlates well with
measured timings on both CPU and GPU.
Model size is defined here as the total number of parameters (network weights) and it relates to memory efficiency.
Larger model sizes tend to yield overfitting (for fixed accuracy).
Architectures closest to the axes origin are both more accurate and more efficient.

The conditional network of Fig.~\ref{fig:Imagenet_CondNet} 
corresponds to the bright green circle in Fig.~\ref{fig:Imagenet_results}.
It achieves a top-5 error of $\sim$13.8\%, identical to that of the VGG11 network (yellow diamond) it is based upon.
However, our conditional network requires less than half the compute (45\%), and almost one-fifth (21\%) of the parameters. 
Our conditional architecture  is the second closest to the origin after GoogLeNet~\cite{Szegedy2014going} 
(in purple).
Both~\cite{Szegedy2014going} and~\cite{Krizhevsky2012imanet} obtain efficiency by sending data to different branches of the network. Although they do not use ``highly branched'' tree structures they can still be thought as special instances of (implicit) conditional networks. GoogLeNet achieves the best results in our joint three-way metric, probably thanks to their use of: 
i) multiple intermediate training losses, 
ii) learnt low-dimensional embeddings, and 
iii) better tuning of the architecture to the specific image dataset. 
Finally, the best accuracy is achieved by~\cite{He2015delving}, but even their most efficient 
model uses ~1.9E+10 flops, and thus falls outside the plot.

\begin{figure}[t]
\centerline{
\includegraphics[width =0.9 \linewidth]{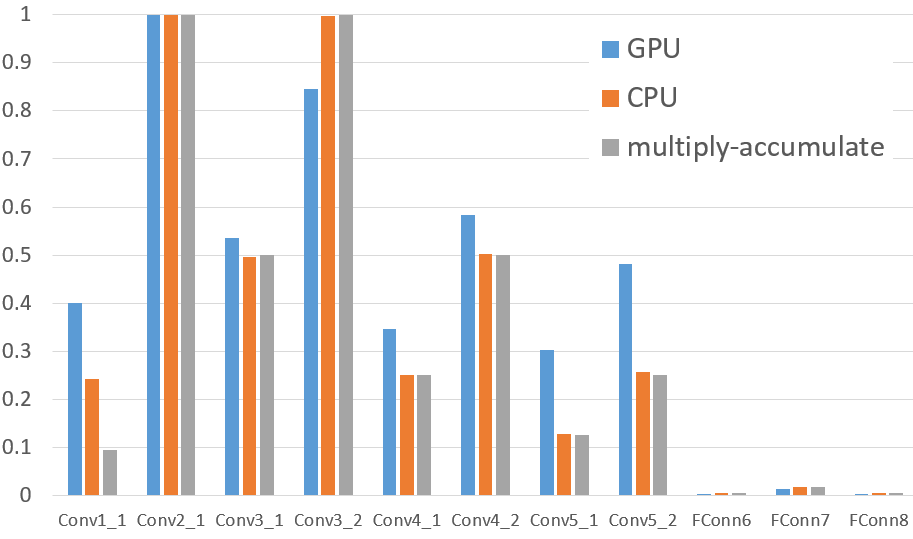}
}
\caption{
{\bf Correlation between predicted layer-wise, test-time compute costs and actual measured timings on CPU and GPU} 
for the conditional architecture in Fig.~\ref{fig:Imagenet_CondNet}. The three histograms have been 
``max normalized'' to aid comparison.
}
\label{fig:Imagenet_timingPlots}
\end{figure}

\noindent{\bf Do fewer operations correspond to faster execution?}
Figure~\ref{fig:Imagenet_timingPlots} reports a layer-wise comparison between the predicted test-time compute cost 
(measured as number of multiply-accumulate operations in the model) and the actual measured timings (both on CPU and GPU) for the network architecture in Fig.~\ref{fig:Imagenet_CondNet}. 
There is a strong correlation between the number of floating-point operations and the actual measured times.
In the GPU case, the correlation is a slightly less strong, due to data moving overheads.
This confirms that, indeed, fewer operations do correspond to faster execution, by roughly the same ratio.
As discussed in Section~\ref{sec:efficiency} this extra speed (compared to conventional CNNs) 
comes from the fact that in branched architectures
successive layers need to run convolutions with smaller shorter kernels, on ever smaller feature maps. 
All architectures tested in our experiments are implemented in the same Caffe framework and enjoy the same two levels of parallelism: i) parallel matrix multiplications (thanks to BLAS\footnote{http://www.netlib.org/blas/}), and ii) data parallelism, thanks to the use of mini-batches. Although highly-branched conditional networks could in theory benefit from {\em model parallelism} (computing different branches on different GPUs, simultaneously), this feature is not yet implemented in Caffe~\cite{Toshev2014deeppose}.

\begin{figure}
\centerline{
\includegraphics[width=1.1\linewidth]{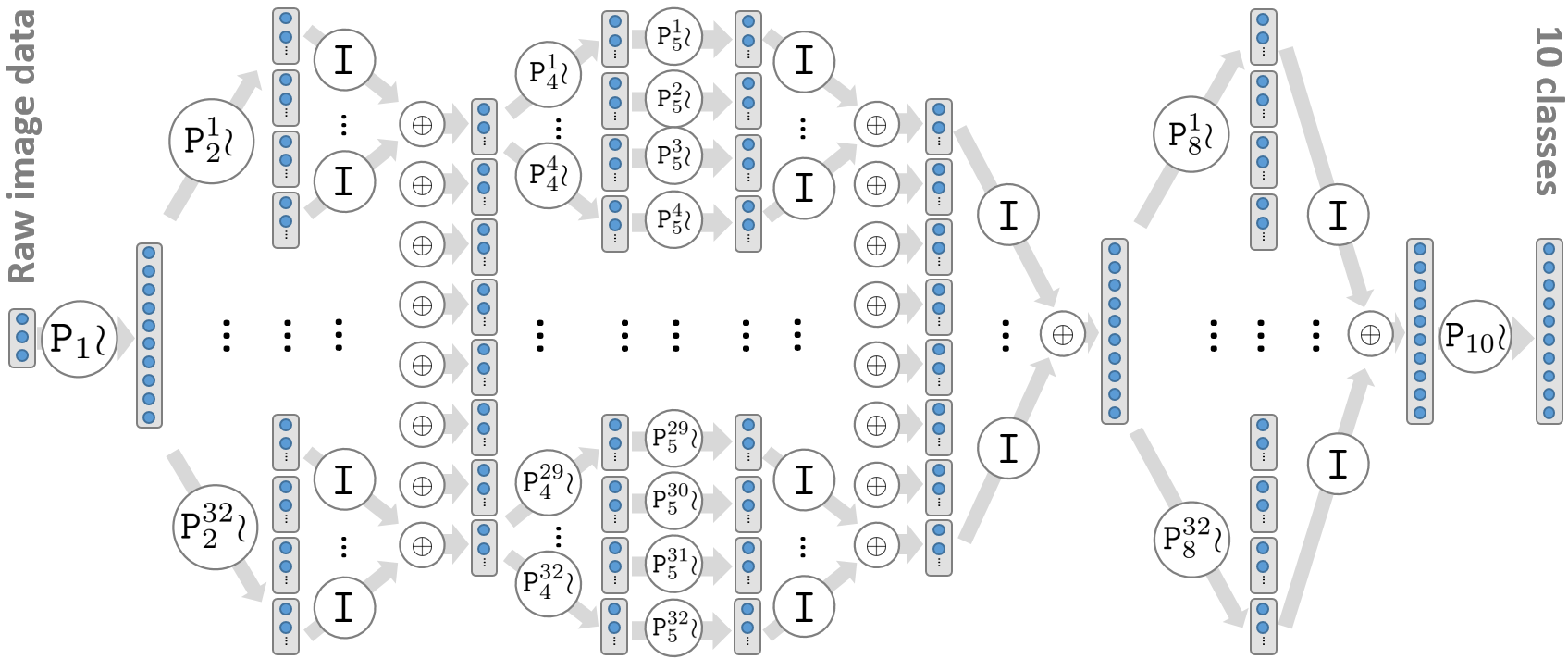}
}
   \caption{{\bf Automatically learned conditional architecture for image classification in CIFAR.} Both structure and parameters of this conditional network have been learned automatically via Bayesian optimization. Best viewed on screen.
}
\label{fig:Cifar_CondNet}
\end{figure}
%

\begin{figure*}[t]
\centerline{
\begin{tabular}{ccc}
\includegraphics[width=0.38\linewidth]{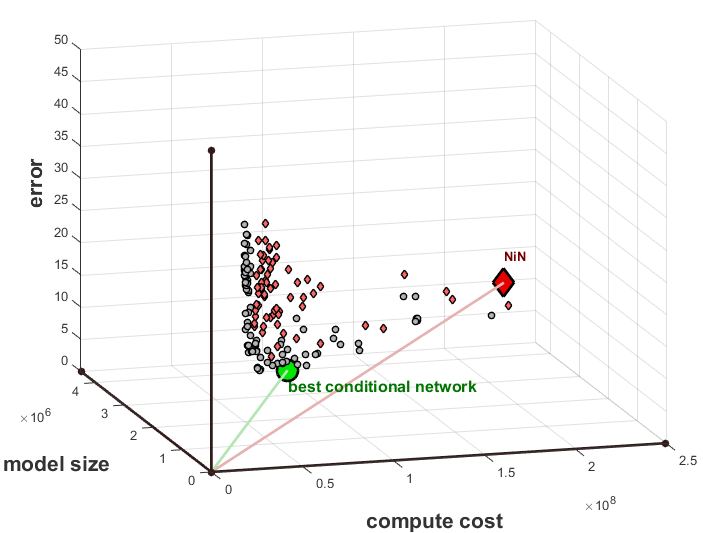}&\hspace{-0.6cm}
\includegraphics[width=0.33\linewidth]{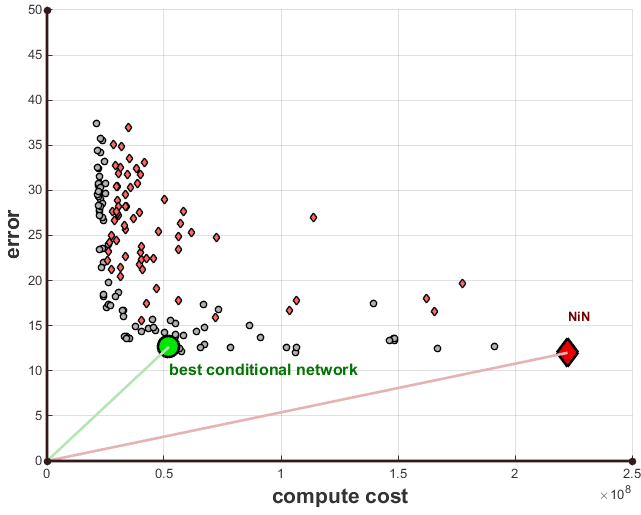}&\hspace{-0.6cm}
\includegraphics[width=0.33\linewidth]{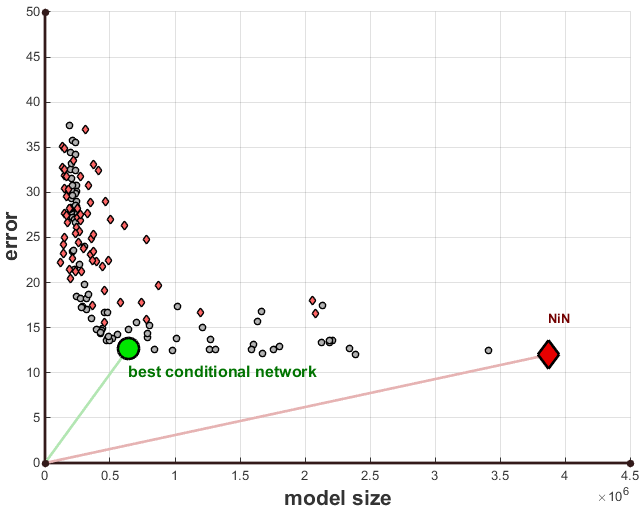}\\
\end{tabular}
}
   \caption{{\bf Comparing network architectures on CIFAR10.} 
Classification error as a function of test-time compute and model size, for various networks, 
validated on the CIFAR10 dataset.
{\bf (left)} A 3D view.
{\bf (middle)} Error \vs compute cost.
{\bf (right)} Error \vs model size.
Our automatically-optimized conditional architecture (green circle) is
{\bf $\sim$5 times faster} and {\bf $\sim$6 times smaller} than NiN, with same accuracy.
}
\label{fig:Cifar_results}
\end{figure*}

\subsection{Comparing Various Architectures on CIFAR}
We further validate our hybrid model on the task of classifying images in the CIFAR10~\cite{CIFAR10} dataset. 
The dataset contains 60K images of 10 classes, typically divided into 50K training images and 10K test images. 
We take the state of the art Network in Network (NiN) model as a reference~\cite{Lin2013NiN}, and we build a conditional version of it.
This time the optimal conditional architecture (in Fig.~\ref{fig:Cifar_CondNet}) is 
constructed {\em automatically}, by using Bayesian search~\cite{Snoek2012} on a parametrized family of architectures.

\noindent{\bf Designing a family of conditional networks.}
The NiN model has a large number (192) of filters in the first convolutional layer, 
representing a sizable amount of the overall compute.\footnote{Most Imagenet networks typically use $64-96$ conv1 filters.}
We build a variant (`NiN-64') that prepends a layer of 64 filters to the NiN model.
While this variant is more complex than NiN, when routed (as described later) it allows us 
to split the larger layers into many routes and increase the efficiency.
By changing the number of routes at each level of the NiN-64 model (from conv2) we can 
generate a whole family of possible conditional architectures.

\noindent{\bf Learning the optimal network architecture.}
Next we search this parametrized space of routed architectures by using Bayesian optimization~\cite{Snoek2012}.
In the optimization we maximized the {\em size-normalized} accuracy $\alpha =\frac{\textrm{classification accuracy}}{\textrm{model size}}$ with respect to the parameters $R_l = 2^i, \left\{i\in \mathbb{N} : 0 \le i \le 5\right\}$, where $R_l$ is the number of nodes at layer $l$ in the conditional network. 
Fig.~\ref{fig:Cifar_CondNet} shows the resulting architecture. It turns out to be a DAG with 10 layers. 

For a fair comparison, we use Bayesian optimization on the NiN architecture too.
We reduce the complexity of the unrouted NiN-64 network by learning a reduction in the number of per-layer filters. 
\ie we maximize $\alpha$ over $F_l = F_\textrm{orig}/2^i, \left\{i\in \mathbb{N} : 0 \le i \le 4\right\}$, where $F_\textrm{orig}$ is the number of filters in layer $l$.
All networks were trained with the same parameters as~\cite{Lin2013NiN}, 
except for using the initialization of~\cite{He2015delving}, 
and a learning schedule of $\gamma_t = \gamma_0(1+\gamma_0\lambda t)^{-1}$, where $\gamma_0,\gamma_t$ and $\lambda$ are the initial learning rate, learning rate at iteration $t$, and weight decay, respectively~\cite{Bottou2012sgdtricks}. 
Training was run for 400 epochs (max), or until the validation 
accuracy had not changed in 10K iterations. 
We split the original training set into 40K training images and 
10K validation images. 
The remaining 10K images are used for testing.

\noindent{\bf Results: accuracy \vs compute \vs size.}
Fig.~\ref{fig:Cifar_results} shows test errors with respect to test-time cost and model size for multiple architectrues.
Diamonds denote unrouted networks and circles denote conditional networks. 
The original NiN is shown in red, and samples of {\em unrouted}, filter-reduced versions explored during the Bayesian optimization are shown in pink.
A sample of 300 {\em conditional} variants are shown as grey circles.
The green circle denotes one such conditional architecture close to the origin of the 
3D space $(test-error, test-cost, model-size)$.
Most of the conditional networks proposed by the optimization are distributed 
near a 3D surface with either low error, low size, low compute cost, or all of them. 
The conditional samples are in average closer to the origin than the unrouted counterparts.
The accuracy of the best conditional network is almost identical to that of the NiN model,
but it is about 5 times faster and 6 times smaller.

\subsection{Conditional Ensembles of CNNs}

A key difference between CNNs and conditional networks is that the latter may include (trainable) data routers. Here we use an {\em explicitly}-routed architecture to create an {\em ensemble} of CNNs where the data traverses only selected, component CNNs (and not necessarily all of them), thus saving computation.

\begin{figure}[t]
\centerline{
\includegraphics[width=1\linewidth]{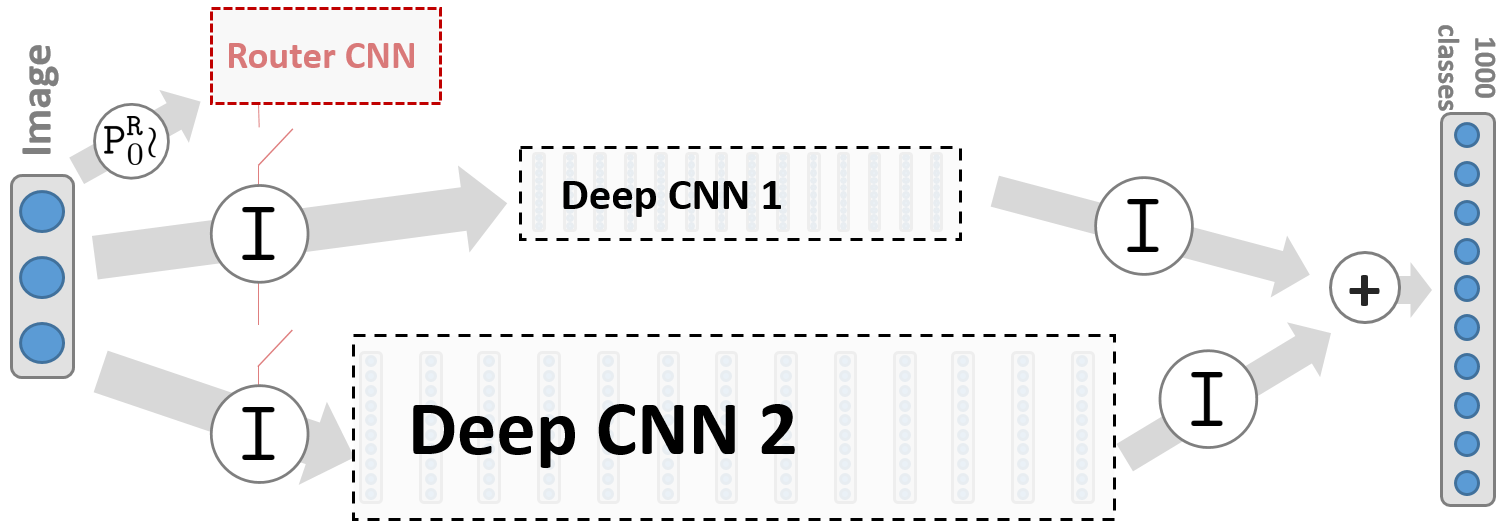}
}
   \caption{{\bf Explicit data routing for conditional ensembles.} 
An explicitly-routed conditional network that mixes existing deep CNNs in a learned, data-dependent fashion.
}
\label{fig:resultsMixing_Network}
\end{figure}

As an example, the branched network in Fig.~\ref{fig:resultsMixing_Network} is applied to the ILSVRC2012 image classification task.
The network has $R=2$ routes, each of which is itself a deep CNN. 
Here, we use GoogLeNet~\cite{Szegedy2014going} as the basis 
of each component route, although other architectures may be used.
Generalizing to $R>2$ is straightforward.
The routes have different compute cost (denoted by different-sized rectangles), arising from differing degrees of test-time oversampling.
We use no oversampling for the first route and 10X oversampling for the second route. 
%
%
The router determines which image should be sent to which route (or both).
The router is trained together with the rest of the network via back-propagation (Section~\ref{sec:training}) to predict the accuracy of each route for each image. 
The router is itself a deep CNN, based on CNN1; This allows computation reuse for extra efficiency.
At test time, a (dynamic) trade off can be made between predicted accuracy and computational cost.

\begin{figure}[t]
\centerline{
\includegraphics[width=1.1\linewidth]{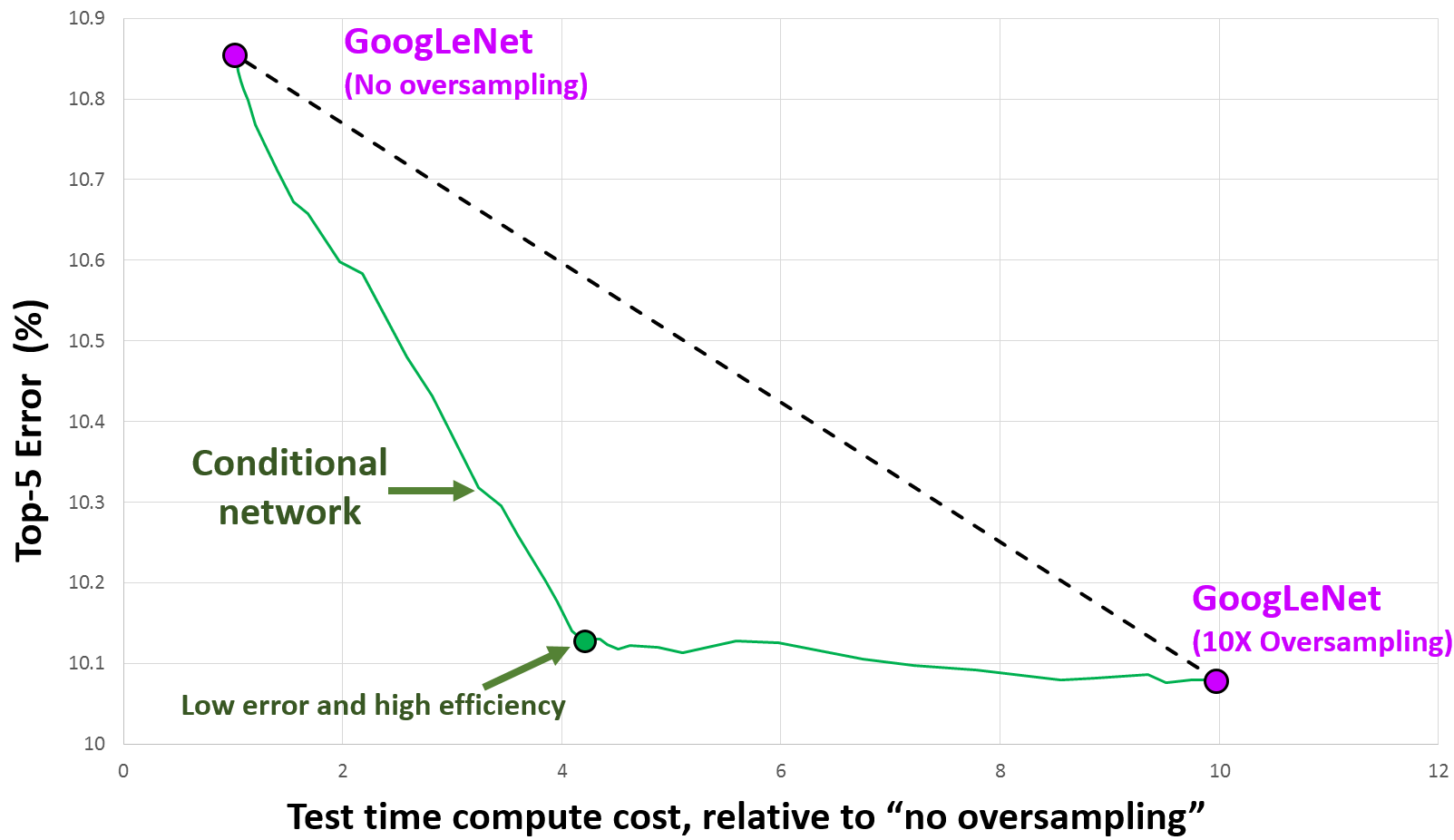}
}
   \caption{{\bf Error-accuracy results for conditional ensembles of CNNs.} 
   Error-accuracy results for the two GoogLeNet base networks are shown in purple.
   The {\em dynamic} error-cost curve for our conditional ensemble is in green.
   In the green circle we achieve same accuracy as the most accurate GoogLeNet with 
   half its cost.
}
\label{fig:resultsMixing_Plot}
\end{figure}

Figure~\ref{fig:resultsMixing_Plot} shows the resulting error-cost curve. 
All costs, including the cost of applying the router are taken into consideration here.
Given our trained conditional network, we use dynamic, multi-way data routing 
(Section~\ref{sec:results_perceptron}) to 
generate a curve in the error-compute space. 
Each point on the curve shows the top-5 error on the 
validation set at a given compute cost, which is an amortized average over the validation set. 
The dashed line corresponds to the trivial error vs. compute trade-off that could be made by selecting one or other base network at random, with a probability chosen so as to achieve a required average compute cost. The fact that the green curve lies significantly below this straight line confirms the much improved
trade-off achieved by the conditional network. 
In the operating point indicated by the green circle we achieve nearly the same accuracy as the $10\times$ oversampled GoogLeNet with less than half its compute cost.
A conventional CNN ensemble
would incur a higher cost since all routes are used for all images.

\section{Discussion and Conclusion}
This paper has investigated similarities and differences between decision trees/forests and convolutional networks. This has led us to introduce a hybrid model (namely conditional network) which can 
be thought both as:
i) trees which have been augmented with representation learning capabilities, and
ii) CNNs which have been augmented with explicit data routers and a rich, branched architecture.

Experiments on image classification have shown that highly branched architectures yield improved accuracy-efficiency trade-off as compared to trees or CNNs.
The desired accuracy-efficiency ratio can be selected {\em at run time}, without the need to train a new network. 
Finally, we have shown how explicit routers can improve the efficiency of {\em ensembles} of CNNs, without loss of accuracy.
We hope these findings will help pave the way to a more systematic exploration
of efficient architectures for deep learning at scale.

{\small
\bibliographystyle{ieee}
\bibliography{references,references_DeepLearning}
}

\end{document}